\documentclass{article}
\usepackage{cite}

\usepackage[pdftex]{graphicx}
\usepackage{subcaption}
\DeclareGraphicsExtensions{.pdf,.jpg,.jpeg,.png}
\usepackage{amsmath}
\usepackage{amssymb}
\usepackage{amsfonts}
\usepackage{algorithmic}
\usepackage{url}

\begin{document}
%
\title{Long-Term Electricity Demand Prediction Using Non-negative Tensor Factorization and Genetic Algorithm-Driven Temporal Modeling
}
%
%
%

\author{Toma Masaki \and
        Kanta Tachibana 
\thanks{T.~Masaki and K.~Tachibana were with the Department
of System Mathematics Science, School of Informatics, Kogakuin University, Japan e-mail:jx21273@g.kogakuin.jp, kanta@cc.kogakuin.ac.jp.}}

\maketitle

\begin{abstract}
This study proposes a novel framework for long-term electricity demand prediction based solely on historical consumption data, without relying on external variables such as temperature or economic indicators. 
The method combines Non-negative Tensor Factorization (NTF) to extract low-dimensional temporal features from multi-way electricity usage data, with a Genetic Algorithm that optimizes the hyperparameters of time series models applied to the latent annual factors.
We model the dataset as a third-order tensor spanning electric utilities, industrial sectors, and years, and apply canonical polyadic decomposition under non-negativity constraints. 
The annual component is forecasted using autoregressive models, with hyperparameter tuning guided by the prediction error or reconstruction accuracy on a validation set. 
Comparative experiments using real-world electricity data from Japan demonstrate that the proposed method achieves lower mean squared error (MSE) than baseline approaches without tensor decomposition or evolutionary optimization. 
Moreover, we find that reducing the model’s degrees of freedom via tensor decomposition improves generalization performance, and that initialization sensitivity in NTF can be mitigated through multiple runs or ensemble strategies.
These findings suggest that the proposed framework offers an interpretable, flexible, and scalable approach to long-term electricity demand prediction and can be extended to other structured time series forecasting tasks.
\end{abstract}


%

\section{Introduction}
%
%
%
%


Forecasting electricity demand is essential for effective energy policy and infrastructure planning.
While overestimation may result in excessive infrastructure investment, underestimation poses an even greater risk—insufficient power supply can lead to widespread energy shortages, disrupting industry, limiting innovation, and ultimately undermining opportunities for economic and societal prosperity \cite{Hong2016, Lago2021}.

Many existing studies incorporate external variables such as meteorological, demographic, or economic indicators to enhance electricity demand estimation \cite{Jie2023, Cao2024}.  
However, such data are often unavailable, unreliable, or inconsistent across regions and time periods.  
An alternative approach is to rely solely on historical electricity consumption data, which is typically more accessible and standardized.  
To make this approach effective—especially for long-term forecasting—it is necessary to adopt advanced methods capable of capturing latent temporal and structural patterns within the consumption data.  
A detailed review of related forecasting methods, tensor-based feature modeling, and hyperparameter optimization techniques is provided in Section~\ref{sec:relatedwork}.

This study proposes a novel framework for long-term electricity demand forecasting based solely on annual consumption data.  
The method combines Non-negative Tensor Factorization for latent feature extraction and Genetic Algorithm-based optimization for time series modeling.  
The objective is to enhance generalization performance by reducing model complexity while maintaining forecasting accuracy.  
In addition, we investigate how sensitive the proposed method is to initialization randomness introduced during the tensor decomposition step.  
This analysis is essential for understanding the robustness and reliability of the framework in practical forecasting settings.

\subsection{Related Work}
\label{sec:relatedwork}

Electricity demand forecasting has been studied extensively using statistical, machine learning, and hybrid approaches.  
Comprehensive reviews by Hong et al.~\cite{Hong2016} and Lago et al.~\cite{Lago2021} summarize advances in probabilistic modeling, neural networks, and ensemble methods.  
Many of these rely on external variables such as temperature, population, or macroeconomic indicators to enhance short- and long-term forecast accuracy \cite{Jie2023, Cao2024}.

Other studies have focused on incorporating sociopolitical or regulatory dimensions, such as policy-driven demand shifts and renewable energy adoption \cite{Sutthichaimethee2019, Haines2013}.  
Time series forecasting using grey-Markov models or seasonal ARIMA has been applied in data-scarce environments \cite{Wang2008, Shiwakoti2023}, showing that consumption-only models can still yield useful predictions.

Short-term load forecasting using ARIMA and neural networks has also been demonstrated on pure consumption data \cite{Nichiforov2017}.  
Fan and Hyndman’s Monash Electricity Forecasting Model (MEFM)~\cite{Fan2012} uses semiparametric additive models with external factors for national-scale long-term forecasting, but its adaptability to variable-scarce scenarios remains unclear.

Tensor decomposition methods—particularly Non-negative Tensor Factorization (NTF)—have been used for pattern extraction and clustering in electricity data \cite{Durand2021, Moriyama2022,Ang2016}, and have also proven effective in domains like image processing and hyperspectral analysis \cite{Jiang2018, Xiong2020}.  
Sparse and incremental extensions of NTF have been developed for recommendation systems and completion tasks \cite{Zhang2022, Biswas2022}, though these do not address time series forecasting.

Xu et al.~\cite{Xu2020} provide a comprehensive survey of tensor-based forecasting.  
Zarnaz et al.~\cite{Zarnaz2021} demonstrate that NTF can improve generalization in financial time series forecasting, highlighting the potential of low-rank latent modeling in structured domains.

In terms of forecasting algorithms, classical ARIMA models typically rely on manually tuned parameters based on domain expertise or statistical criteria \cite{Fattah2018}.  
Genetic Algorithms have been adopted to automate parameter search in hybrid models and epidemic prediction tasks \cite{Panda2024, Zhang2017}.  
Their flexibility makes them well-suited for non-convex, high-dimensional search spaces like those arising in time series modeling.

While deep learning methods have become increasingly prominent in short-term forecasting \cite{Duan2019}, their application to long-term national-scale prediction remains limited due to their data and compute requirements, and lack of interpretability.

In this context, our work integrates NTF and Genetic Algorithm-based time series modeling to support interpretable, scalable, and data-efficient long-term electricity demand prediction, without requiring external covariates.

\subsection{Theoretical Background}

\subsubsection{Non-negative Tensor Factorization (NTF)}

Let \( \mathcal{X} \in \mathbb{R}^{I \times J \times K} \) be a third-order non-negative tensor representing electricity consumption data across electric utilities, industries, and years. NTF aims to approximate \( \mathcal{X} \) by decomposing it into three factor matrices:
\[
\mathcal{X} \approx \sum_{r=1}^R \mathbf{a}_r \circ \mathbf{b}_r \circ \mathbf{c}_r
\]
where \( \mathbf{A} = [\mathbf{a}_1, \dots, \mathbf{a}_R] \in \mathbb{R}^{I \times R} \), \( \mathbf{B} = [\mathbf{b}_1, \dots, \mathbf{b}_R] \in \mathbb{R}^{J \times R} \), and \( \mathbf{C} = [\mathbf{c}_1, \dots, \mathbf{c}_R] \in \mathbb{R}^{K \times R} \) are non-negative factor matrices, and \( \circ \) denotes the outer product.
We denote by \( X_{ijk} \) the \((i,j,k)\)-th element of \( \mathcal{X} \), and by \( \tilde{X}_{ijk} \) the corresponding element of the approximated tensor \( \tilde{\mathcal{X}} \).  
Each factor vector \( \mathbf{a}_r \in \mathbb{R}^{I} \), \( \mathbf{b}_r \in \mathbb{R}^{J} \), and \( \mathbf{c}_r \in \mathbb{R}^{K} \) contains non-negative elements, where \( a_{ir} \), \( b_{jr} \), and \( c_{kr} \) denote the \( i \)-th, \( j \)-th, and \( k \)-th elements of \( \mathbf{a}_r \), \( \mathbf{b}_r \), and \( \mathbf{c}_r \), respectively.

The update rules for multiplicative optimization are:
\begin{align}
a_{ir} &\leftarrow a_{ir} \frac{\sum_{j,k} \left(X_{ijk} / \tilde{X}_{ijk} \right)\cdot b_{jr} c_{kr}}{\sum_{j,k} b_{jr} c_{kr}}, \\
b_{jr} &\leftarrow b_{jr} \frac{\sum_{i,k} \left(X_{ijk} / \tilde{X}_{ijk} \right)\cdot a_{ir} c_{kr}}{\sum_{i,k} a_{ir} c_{kr}}, \\
c_{kr} &\leftarrow c_{kr} \frac{\sum_{i,j} \left(X_{ijk} / \tilde{X}_{ijk} \right)\cdot a_{ir} b_{jr}}{\sum_{i,j} a_{ir} b_{jr}}, \\
\tilde{X}_{ijk} &= \sum_{r=1}^R a_{ir} b_{jr} c_{kr}.
\end{align}
Reconstruction accuracy $\alpha$ is measured by the Frobenius norm:
\[
\alpha = 1 - \frac{\|\mathcal{X} - \tilde{\mathcal{X}}\|_F}{\|\mathcal{X}\|_F}
\]

\subsubsection{Genetic Algorithm for Time Series Modeling}

We use a Genetic Algorithm (GA) to optimize the hyperparameters \( (p_r, d_r, q_r) \) for each time series model applied to the latent temporal factors extracted by NTF. The GA evolves individuals representing sequences:

\[
[p_1, d_1, q_1, \ldots, p_R, d_R, q_R]
\]

Each individual’s fitness is defined as the negative mean squared error (MSE) between the predicted and actual electricity demand during the validation period.

Selection is performed using tournament selection, and crossover/mutation are applied to promote diversity. This process enables adaptive modeling of latent time series while avoiding overfitting and underfitting issues associated with fixed ARIMA parameters.

\section{Methodologies}

\subsection{Data Description and Tensor Construction}

The dataset used in this study is derived from publicly available electricity consumption records published by the Federation of Electric Power Companies of Japan. It includes annual consumption data from 1963 to 2015, recorded across ten regional electric utilities and detailed industry classifications.

The industry labels consist of 36 categories organized hierarchically. These are aggregated into 10 major industrial sectors: Mining, Manufacturing, Chemical Industry, Petroleum and Coal Products Manufacturing, Rubber Product Manufacturing, Ceramic and Stone Product Manufacturing, Steel Industry, Non-ferrous Metal Manufacturing, Machinery and Equipment Manufacturing 1, and Machinery and Equipment Manufacturing 2.

The full dataset is modeled as a third-order non-negative tensor \( \mathcal{X} \in \mathbb{R}^{I \times J \times K} \), where \( I = 10 \) (electric utilities), \( J = 10 \) (industries), and \( K = 53 \) (years). This tensor serves as the input to the Non-negative Tensor Factorization (NTF) process described in Section~1.2.1.

To evaluate forecasting performance, the dataset is split into training, validation, and test sets along the year mode. Two test cases are constructed:

\begin{itemize}
  \item \textbf{Case A2010:} Training data from 1963–2003, validation data from 2004–2009, test data from 2010–2015.
  \item \textbf{Case A2000:} Training data from 1963–1993, validation data from 1994–1999, test data from 2000–2005.
\end{itemize}

\subsection{Tensor Decomposition and Rank Selection}

The training tensor \( \mathcal{X}_{\text{train}} \) is decomposed using the canonical polyadic (CP) NTF method. The number of update iterations is fixed at 100, and the rank \( R \) is selected by evaluating reconstruction accuracy (as defined in Section~1.2.1) across different candidate ranks.

\begin{figure}[htbp!]
\centering
\includegraphics[width=\textwidth]{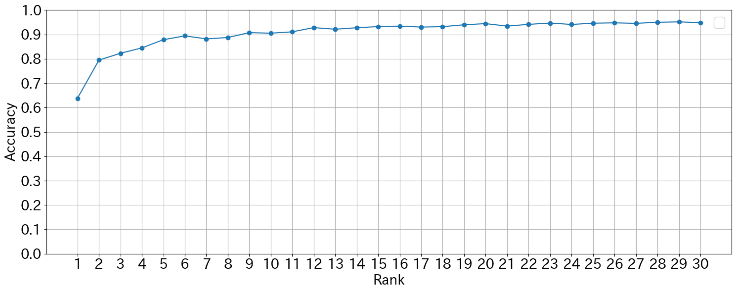}
\caption{Relationship between CP decomposition rank and reconstruction accuracy.}
\label{fig:rank_reconstruction}
\end{figure}

Figure~\ref{fig:rank_reconstruction} shows the reconstruction accuracy for each rank. Based on a trade-off between reconstruction quality and computational cost during the Genetic Algorithm (GA) optimization, we choose \( R = 8 \) for all experiments.

\subsection{Time Series Forecasting on Latent Factors}

Following tensor decomposition, we apply time series forecasting to each column of the annual factor matrix \( \mathbf{C} \in \mathbb{R}^{K \times R} \). Each time series \( \mathbf{c}_r \) is modeled using an ARIMA(\(p_r, d_r, q_r\)) process, where hyperparameters are optimized via a Genetic Algorithm as outlined in Section~1.2.2.

The predictive annual factor matrix for the validation period is denoted as:
\[
\hat{\mathbf{C}} = [\hat{\mathbf{c}}_1, \hat{\mathbf{c}}_2, \dots, \hat{\mathbf{c}}_R]
\]

The reconstructed tensor for the validation period is then:
\[
\hat{\mathcal{X}} = \sum_{r=1}^{R} \mathbf{a}_r \circ \mathbf{b}_r \circ \hat{\mathbf{c}}_r
\]

The GA selects the best set of ARIMA parameters by minimizing the mean squared error (MSE) between the sum of predicted consumption and the ground-truth validation tensor.

\subsection{Baseline Methods for Comparison}

To evaluate the effectiveness of our NTF+GA-based method, we define two baseline methods:

\begin{itemize}
  \item \textbf{w/o NTF:} Apply ARIMA(3,2,3) directly to each of the 100 time series corresponding to (utility, industry) combinations, bypassing tensor decomposition.
  \item \textbf{w/o GA:} Apply fixed ARIMA(3,2,3) to each latent annual factor obtained from NTF without performing GA-based hyperparameter search.
\end{itemize}

These settings allow us to isolate the impact of NTF and GA independently.

\subsection{Evaluation Protocol and Test Case Patterns}

We define four evaluation patterns depending on the test case and fitness criterion used during GA optimization:

\begin{itemize}
  \item \textbf{Pattern A2010:} GA optimized using MSE on total consumption (Case A2010).
  \item \textbf{Pattern B2010:} GA optimized using reconstruction accuracy (Frobenius norm) (Case A2010).
  \item \textbf{Pattern A2000:} GA optimized using MSE (Case A2000).
  \item \textbf{Pattern B2000:} GA optimized using reconstruction accuracy (Case A2000).
\end{itemize}

The ARIMA hyperparameters used in fixed settings are shown in Table~\ref{tab:arima_params}.

\begin{table}[htbp!]
\centering
\caption{ARIMA hyperparameters used in baseline methods.}
\begin{tabular}{l|ccc}
\hline
Pattern & \( p \) & \( d \) & \( q \) \\
\hline
A2010 & 3 & 2 & 3 \\
B2010 & 2 & 1 & 3 \\
A2000 & 3 & 2 & 2 \\
B2000 & 2 & 1 & 2 \\
\hline
\end{tabular}
\label{tab:arima_params}
\end{table}

\section{Results}

\subsection{Prediction Accuracy}

\begin{figure}[htbp!]
\centering
\includegraphics[width=\linewidth]{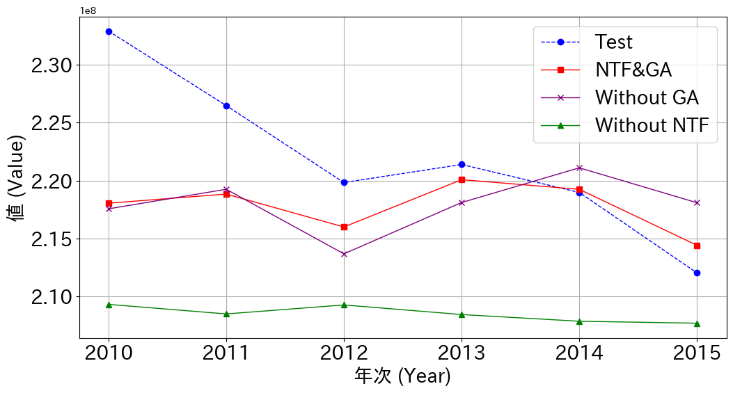}
\caption{Prediction results for Pattern A2010.}
\label{fig:predict_A2010}
\end{figure}

\begin{figure}[htbp!]
\centering
\includegraphics[width=\linewidth]{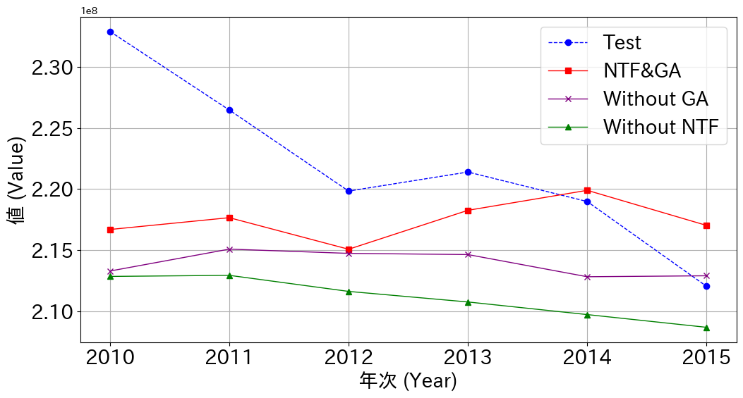}
\caption{Prediction results for Pattern B2010.}
\label{fig:predict_B2010}
\end{figure}

\begin{figure}[htbp!]
\centering
\includegraphics[width=\linewidth]{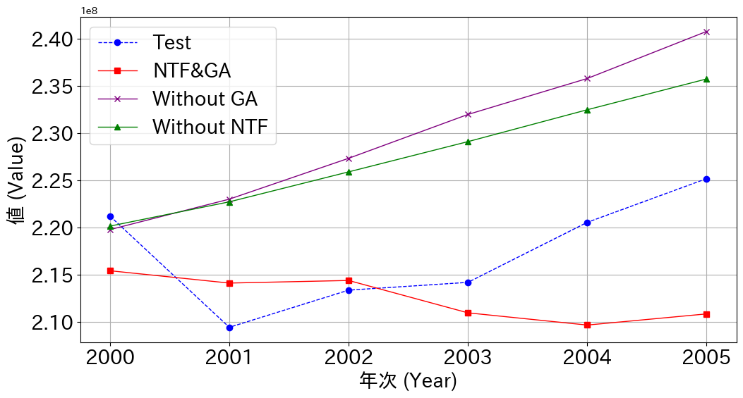}
\caption{Prediction results for Pattern A2000.}
\label{fig:predict_A2000}
\end{figure}

\begin{figure}[htbp!]
\centering
\includegraphics[width=\linewidth]{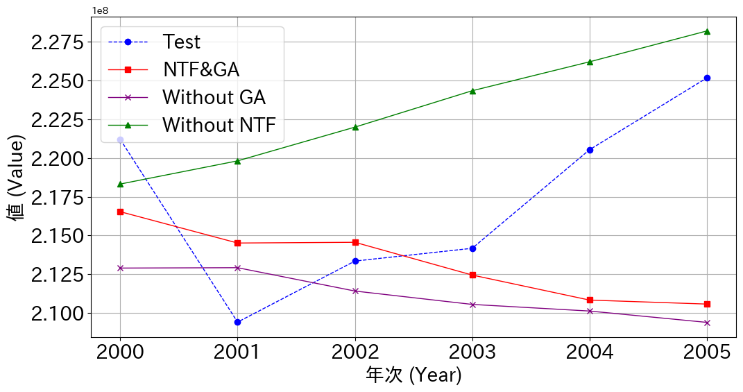}
\caption{Prediction results for Pattern B2000.}
\label{fig:predict_B2000}
\end{figure}

We evaluate the proposed method and the two baseline methods (w/o NTF, w/o GA) across four experimental patterns described in Section~2.5. Tables~\ref{tab:result_A2010}–\ref{tab:result_B2000} summarize the MSE and reconstruction accuracy for each method. Figures~\ref{fig:predict_A2010}–\ref{fig:predict_B2000} illustrate the annual total electricity demand predicted by each method compared with the ground truth.

\begin{table}[htbp!]
\centering
\caption{MSE and reconstruction accuracy for Pattern A2010.}
\begin{tabular}{lcc}
\hline
Method & MSE (\( \times 10^{13} \)) & Reconstruction \\
\hline
Proposed & 5.022 & 0.811 \\
w/o GA   & 6.292 & 0.819 \\
w/o NTF  & 21.70 & 0.816 \\
\hline
\end{tabular}
\label{tab:result_A2010}
\end{table}

\begin{table}[htbp!]
\centering
\caption{MSE and reconstruction accuracy for Pattern B2010.}
\begin{tabular}{lcc}
\hline
Method & MSE (\( \times 10^{13} \)) & Reconstruction \\
\hline
Proposed & 6.655 & 0.825 \\
w/o GA   & 10.43 & 0.818 \\
w/o NTF  & 14.40 & 0.883 \\
\hline
\end{tabular}
\label{tab:result_B2010}
\end{table}

\begin{table}[htbp!]
\centering
\caption{MSE and reconstruction accuracy for Pattern A2000.}
\begin{tabular}{lcc}
\hline
Method & MSE (\( \times 10^{13} \)) & Reconstruction \\
\hline
Proposed & 6.524 & 0.799 \\
w/o GA   & 19.60 & 0.795 \\
w/o NTF  & 13.54 & 0.892 \\
\hline
\end{tabular}
\label{tab:result_A2000}
\end{table}

\begin{table}[htbp!]
\centering
\caption{MSE and reconstruction accuracy for Pattern B2000.}
\begin{tabular}{lcc}
\hline
Method & MSE (\( \times 10^{13} \)) & Reconstruction \\
\hline
Proposed & 5.995 & 0.818 \\
w/o GA   & 7.595 & 0.842 \\
w/o NTF  & 5.606 & 0.913 \\
\hline
\end{tabular}
\label{tab:result_B2000}
\end{table}

\subsection{Influence of Random Seed Initialization}

\begin{figure}[htbp!]
\centering
\includegraphics[width=\linewidth]{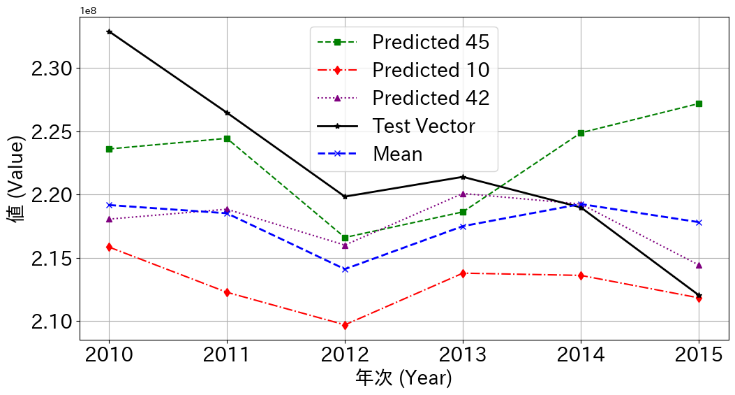}
\caption{Prediction results with different seeds for Pattern A2010.}
\label{fig:randomseed_A2010}
\end{figure}

\begin{figure}[htbp!]
\centering
\includegraphics[width=\linewidth]{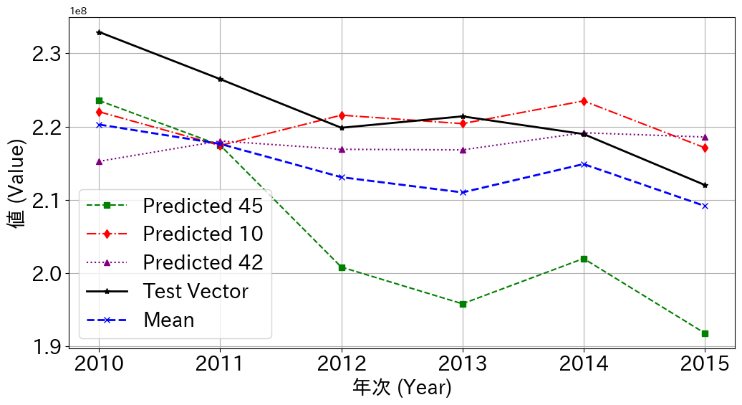}
\caption{Prediction results with different seeds for Pattern B2010.}
\label{fig:randomseed_B2010}
\end{figure}

\begin{figure}[htbp!]
\centering
\includegraphics[width=\linewidth]{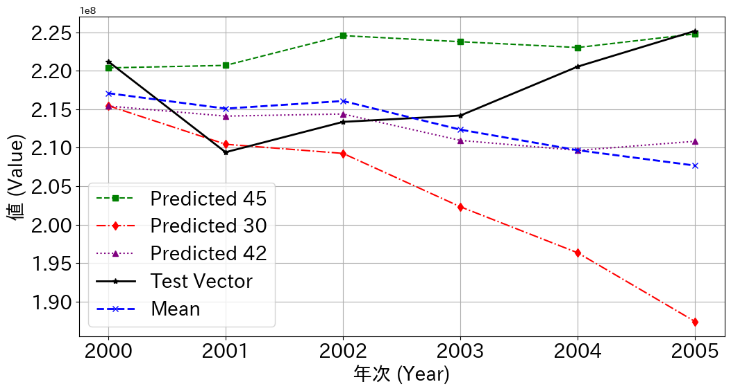}
\caption{Prediction results with different seeds for Pattern A2000.}
\label{fig:randomseed_A2000}
\end{figure}

\begin{figure}[htbp!]
\centering
\includegraphics[width=\linewidth]{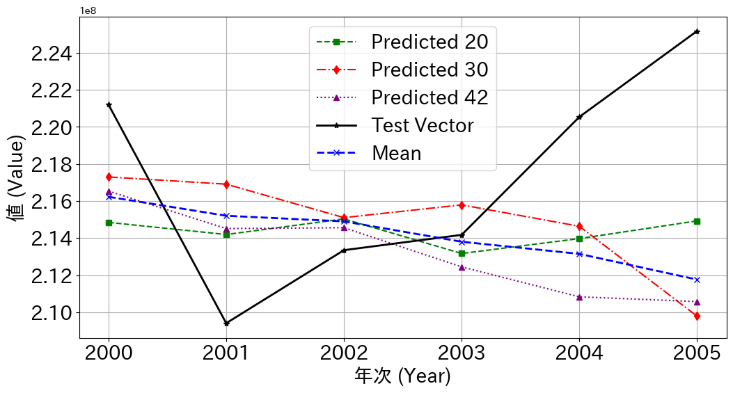}
\caption{Prediction results with different seeds for Pattern B2000.}
\label{fig:randomseed_B2000}
\end{figure}

To assess the stability of the proposed method, we examine how reconstruction accuracy varies with different random seed values used during NTF initialization. Tables~\ref{tab:seed_A2010}–\ref{tab:seed_B2000} show reconstruction accuracy on validation data for five different seeds: 10, 20, 30, 42, and 45. The corresponding prediction results, averaged over the top three reconstruction runs, are plotted in Figures~\ref{fig:randomseed_A2010}–\ref{fig:randomseed_B2000}.

\begin{table}[htbp!]
\centering
\caption{Reconstruction accuracy on validation data for Pattern A2010 (by seed).}
\begin{tabular}{cc}
\hline
Seed & Reconstruction \\
\hline
10  & 0.904 \\
45  & 0.890 \\
42  & 0.888 \\
\hline
30  & 0.876 \\
20  & 0.862 \\
\hline
\end{tabular}
\label{tab:seed_A2010}
\end{table}

\begin{table}[htbp!]
\centering
\caption{Reconstruction accuracy on validation data for Pattern B2010 (by seed).}
\begin{tabular}{cc}
\hline
Seed & Reconstruction \\
\hline
10  & 0.904 \\
45  & 0.890 \\
42  & 0.888 \\
\hline
30  & 0.875 \\
20  & 0.862 \\
\hline
\end{tabular}
\label{tab:seed_B2010}
\end{table}

\begin{table}[htbp!]
\centering
\caption{Reconstruction accuracy on validation data for Pattern A2000 (by seed).}
\begin{tabular}{cc}
\hline
Seed & Reconstruction \\
\hline
42  & 0.886 \\
30  & 0.878 \\
45  & 0.865 \\
\hline
20  & 0.623 \\
10  & 0.601 \\
\hline
\end{tabular}
\label{tab:seed_A2000}
\end{table}

\begin{table}[htbp!]
\centering
\caption{Reconstruction accuracy on validation data for Pattern B2000 (by seed).}
\begin{tabular}{cc}
\hline
Seed & Reconstruction \\
\hline
30  & 0.897 \\
20  & 0.896 \\
42  & 0.895 \\
\hline
10  & 0.879 \\
45  & 0.868 \\
\hline
\end{tabular}
\label{tab:seed_B2000}
\end{table}

\section{Discussion}

This section discusses key findings from the experimental results and analyzes the implications in terms of model generalization, reconstruction accuracy, and sensitivity to initialization.

\subsection{Impact of Fitness Criteria on Forecast Accuracy}

Comparing Patterns A2010 and B2010, where the only difference lies in the GA fitness function (MSE versus reconstruction accuracy), we observe contrasting behavior. As shown in Tables~\ref{tab:result_A2010} and~\ref{tab:result_B2010}, optimizing for reconstruction accuracy leads to higher structural fidelity (0.825 vs. 0.811), while optimizing for MSE yields better forecasting performance in terms of true error (lower MSE).

Interestingly, in Patterns A2000 and B2000 (Tables~\ref{tab:result_A2000} and~\ref{tab:result_B2000}), both MSE and reconstruction accuracy improve when the GA is optimized for reconstruction accuracy. This suggests that the optimal fitness criterion may depend on the data range, underlying consumption trends, or the degree of seasonality and noise.

\subsection{Effect of Model Complexity and Degrees of Freedom}

To understand the generalization ability of each model, we analyze the total number of degrees of freedom (DOF) associated with each configuration. Table~\ref{tab:degrees_of_freedom} summarizes the DOF for each experimental pattern in the proposed method, computed as:

\[
\text{DOF} = \sum_{r=1}^R (p_r + q_r)
\]

We find that configurations with fewer DOF (e.g., B2010 and B2000) generally result in better reconstruction accuracy and sometimes better MSE, supporting the hypothesis that model simplicity contributes to generalization.

\begin{table}[htbp!]
\centering
\caption{Total degrees of freedom in the proposed method.}
\begin{tabular}{lccc}
\hline
Pattern & \( \sum p_r \) & \( \sum q_r \) & DOF = \( \sum (p_r + q_r) \) \\
\hline
A2010 & 25 & 21 & 46 \\
B2010 & 16 & 14 & 30 \\
A2000 & 24 & 23 & 47 \\
B2000 & 16 & 13 & 29 \\
\hline
\end{tabular}
\label{tab:degrees_of_freedom}
\end{table}

Notably, the w/o NTF baseline exhibits the highest reconstruction accuracy in several patterns. However, this is likely due to its increased effective DOF per tensor element, as ARIMA is applied independently to each utility–industry pair. In contrast, our method enforces structural constraints through NTF, which may sacrifice reconstruction accuracy but improves generalization by capturing shared latent patterns.

\subsection{Sensitivity to Initialization}

As shown in Tables~\ref{tab:seed_A2010}–\ref{tab:seed_B2000}, the NTF-based method exhibits variability in reconstruction accuracy depending on the random seed used during initialization. This is consistent with known behavior of non-convex matrix/tensor factorization methods, where local minima can lead to different decompositions.

Nonetheless, Figures~\ref{fig:randomseed_A2000} and~\ref{fig:randomseed_B2000} show that averaging over top-performing initializations provides stable predictions. This suggests that incorporating multiple initialization trials or ensemble methods could further enhance robustness.

\subsection{Interpretability and Scalability}

By reducing the dimensionality of the original data through NTF, the proposed method achieves interpretable factor matrices that reflect underlying temporal trends. The use of GA further automates the selection of appropriate time series model complexity for each factor, allowing for adaptive modeling without manual tuning.

Moreover, the framework is scalable and extensible. It can be adapted to include additional modes (e.g., geographic regions, customer classes) or applied to other domains such as water or gas demand, provided a structured tensor representation is available.

\section{Conclusion}

This study presented a novel framework for long-term electricity demand forecasting using only historical consumption data. By leveraging Non-negative Tensor Factorization (NTF) to extract latent temporal features and optimizing time series model parameters through a Genetic Algorithm (GA), the proposed method achieves both interpretability and competitive prediction accuracy.

Experimental evaluations on real-world electricity consumption data from Japan demonstrated that the proposed approach consistently outperforms baseline methods in terms of mean squared error (MSE), particularly when model complexity is effectively reduced via latent factor modeling. The use of GA for adaptive ARIMA parameter tuning was shown to be effective across multiple forecasting scenarios.

Our analysis further revealed that minimizing reconstruction error during GA optimization can improve generalization, especially when accompanied by a reduction in the total degrees of freedom. Moreover, while NTF introduces sensitivity to initialization, the impact on predictive performance can be mitigated through multiple trials or ensemble strategies.

The proposed framework is domain-independent and can be extended to other utility forecasting tasks, such as water or gas consumption, provided the data can be represented as structured tensors. Future work includes incorporating adaptive rank selection, robust initialization strategies, and expanding the method to support exogenous variables or probabilistic forecasting outputs.


%








\bibliographystyle{IEEEtran}
\bibliography{./refs.bib}
%

%








\end{document}